\documentclass[10pt,twocolumn,letterpaper]{article}

\usepackage[pagenumbers]{cvpr} 

\usepackage{graphicx}
\usepackage{amsmath}
\usepackage{amssymb}
\usepackage{booktabs}
\usepackage{comment}

\usepackage[pagebackref,breaklinks,colorlinks]{hyperref}

\usepackage{algorithm}
\usepackage{algpseudocode}
\usepackage{float}
\usepackage{color}
\usepackage{colortbl}
\usepackage{rotating}
\usepackage{times}
\usepackage{epsfig}
\usepackage{enumitem}
\usepackage{makecell}
\usepackage{lineno}
\usepackage{tabularx}
\usepackage{xcolor}
\usepackage{multirow}
\usepackage{graphicx}
\usepackage{hhline}
\usepackage{textcomp,booktabs}
\usepackage{caption}
\usepackage{stmaryrd}
\usepackage[mathscr]{eucal}
\usepackage{lineno}
\usepackage[export]{adjustbox}
\usepackage{comment}
\usepackage{multicol,lipsum}
\usepackage{amsfonts}

\def\1{mathbb{1}}

\def\x{{\bf x}}

\def\W{{\bf W}}
\def\w{{\bf w}}
\def\0{{\bf 0}}
\def\1{{\bf 1}}

\def\RB{{\mathbb R}}

\def\tha{\mbox{\boldmath$\theta$\unboldmath}}
\def\Tha{\mbox{\boldmath$\Theta$\unboldmath}}

\definecolor{purple}{rgb}{0.56,0.27,0.68}
\definecolor{red}{rgb}{0.95,0.4,0.4}
\definecolor{purered}{rgb}{1,0,0}
\definecolor{blue}{rgb}{0.4,0.4,0.95}
\definecolor{darkblue}{rgb}{0,0,0.8}
\definecolor{grey}{rgb}{0.6,0.6,0.6}
\definecolor{col1}{RGB}{232, 161, 148}
\definecolor{col2}{RGB}{148, 187, 232}
\definecolor{lightgrey}{rgb}{0.85,0.85,0.85}
\definecolor{lightlightgrey}{rgb}{0.9,0.9,0.9}
\definecolor{verylightBG}{rgb}{0.9,0.99,0.99}
\definecolor{darkgreen}{rgb}{0.3, 0.75, 0.3}

\graphicspath{{./figures/}}   

\newcommand\orange[1]{\textcolor{orange}{#1}}

\newcommand\darkblue[1]{\textcolor{darkblue}{#1}}

\usepackage[capitalize]{cleveref}
\crefname{section}{Sec.}{Secs.}
\Crefname{section}{Section}{Sections}
\Crefname{table}{Table}{Tables}
\crefname{table}{Tab.}{Tabs.}

\def\cvprPaperID{0020} 
\def\confName{CVPR}
\def\confYear{2022}

\begin{document}

\title{Long-Tailed Recognition via Weight Balancing}

\author{
Shaden Alshammari\textsuperscript{$\ast$} \quad\quad
Yu-Xiong Wang\textsuperscript{$\natural$} \quad\quad
Deva Ramanan\textsuperscript{$\sharp$,$\dag$} \quad\quad
Shu Kong\textsuperscript{$\sharp$}    \\
\textsuperscript{$\ast$}MIT   \quad\quad  
\textsuperscript{$\natural$}UIUC  \quad\quad   \textsuperscript{$\dag$}Argo AI  \quad\quad
\textsuperscript{$\sharp$}CMU  \\
{\small {\tt shaden@mit.edu} \quad {\tt yxw@illinois.edu} \quad \{\tt deva, shuk\}@andrew.cmu.edu} 
\\
{\small \url{https://github.com/ShadeAlsha/LTR-weight-balancing}}
\vspace{-3.5mm}
}


\twocolumn[
{%
\vspace{-1em}
\maketitle
\vspace{-2mm}
\centering
\begin{minipage}[t]{\textwidth}
\centering
{\small {\bf (a)} per-class classification accuracy vs. class cardinality on CIFAR100-LT (imalance factor 100)}  \\
\includegraphics[width=0.90\textwidth, clip, trim={3cm 0cm 1.5cm 1.2cm}]{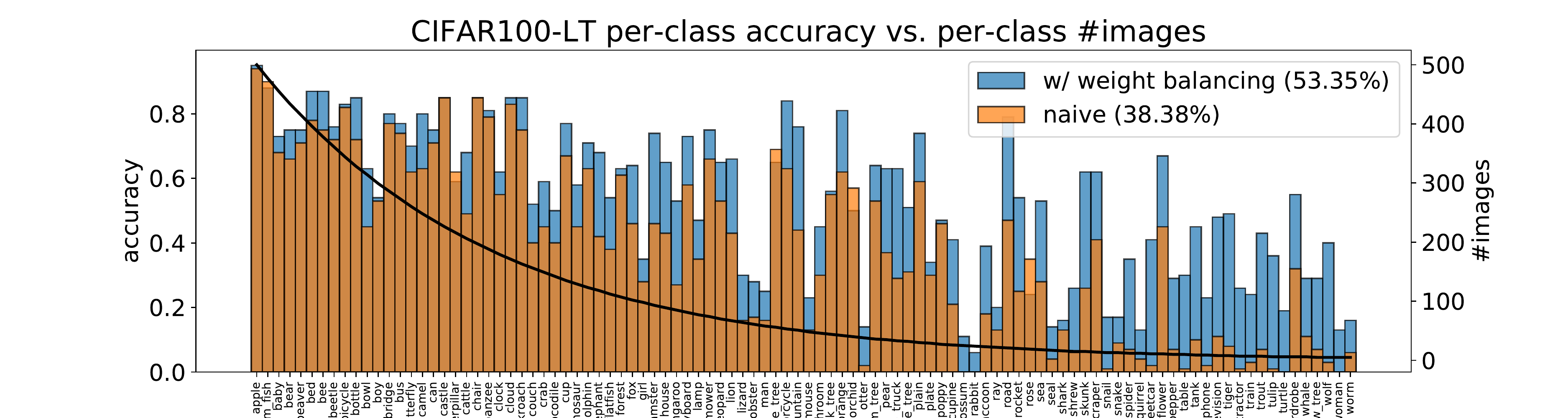}
\end{minipage} \\
\vspace{2mm}
{\small {\bf (b)} norms of per-class weights from the learned classifier vs. class cardinality}  \\
\begin{minipage}[t]{1\textwidth}
\centering 
\includegraphics[width=0.80\textwidth, clip, trim={9cm 0cm 8cm 0cm}]{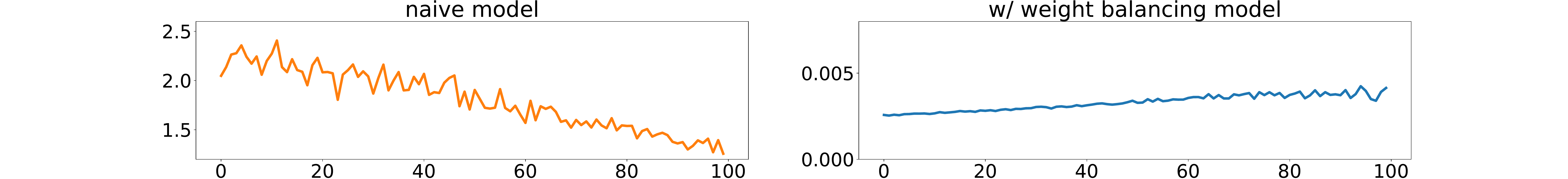}
\end{minipage} 
\vspace{-6.5mm}
\captionof{figure}{\small
Long-tailed recognition (LTR) requires training on long-tailed class distributed data (black curve in {\bf (a)}). {\bf (a)} Networks naively trained on such data are biased toward common classes in terms of higher accuracy (\orange{orange bars}). {\bf (b)}-left plots the L2 norms of per-class weights in the naive classifier. Clearly, classifiers of common classes have ``artificially'' larger norms than rare ones, because they are trained with more data. This can result in over-predictions of common classes, or alternatively, under-predictions of rare classes. This observation motivates us to balance norms via {\em parameter regularization}. To do so, we explore simple weight balancing techniques including L2-normalization, weight decay, and the MaxNorm constraint. We find that applying the latter two results in class weights to be far more balanced ({\bf (b)}-right), allowing rare classes to have a ``fighting chance'' when competing with common classes.  Our model boosts overall accuracy to 53.35\% (\darkblue{blue bars} in {\bf (a)}),  significantly higher than the naive model (38.38\%) and prior art, e.g., RIDE (49.1\%)~\cite{wang2020long}, ACE (49.6\%)~\cite{cai2021ace}, and PaCo (52.0\%)~\cite{cui2021parametric}. 
Results are from experiments (Table~\ref{tab:CIFAR100-ablation}) on CIFAR100-LT with an imbalance factor 100~\cite{cao2019learning}. 
}
\label{fig:splashing}
\vspace{5mm}
}
]

\begin{abstract} \vspace{-2mm}
In the real open world, data tends to follow long-tailed class distributions, motivating the well-studied long-tailed recognition (LTR) problem. Naive training produces models that are biased toward common classes in terms of higher accuracy. The key to addressing LTR is to balance various aspects including data distribution, training losses, and gradients in learning. We explore an orthogonal direction, {\bf weight balancing}, motivated by the empirical observation that the naively trained classifier has ``artificially'' larger weights in norm for common classes (because there exists abundant data to train them, unlike the rare classes). We investigate three techniques to balance weights, L2-normalization, weight decay, and MaxNorm. We first point out that L2-normalization ``perfectly'' balances per-class weights to be unit norm, but such a hard constraint might prevent classes from learning better classifiers. In contrast, weight decay penalizes larger weights more heavily and so learns small balanced weights; the MaxNorm constraint encourages growing small weights within a norm ball but caps all the weights by the radius. Our extensive study shows that both help learn balanced weights and greatly improve the LTR accuracy. Surprisingly, weight decay, although underexplored in LTR, significantly improves over prior work. Therefore, we adopt a two-stage training paradigm and propose a simple approach to LTR: (1) learning features using the cross-entropy loss by tuning weight decay, and (2) learning classifiers using class-balanced loss by tuning weight decay and MaxNorm. Our approach achieves the state-of-the-art accuracy on five standard benchmarks, serving as a future baseline for long-tailed recognition.
\end{abstract}

\section{Introduction}
\label{sec:intro}


In the real open world, data tends to follow long-tailed  distributions~\cite{reed2001pareto, buda2018systematic, zhao2021camera, zhang2021deep}.
Through the lens of classification, 
this means that the number of per-class data, or class cardinality, is heavily imbalanced~\cite{gupta2019lvis, van2018inaturalist}.
Numerous applications emphasize the rare classes.
For example, 
autonomous vehicles should recognize not only common objects such as cars and pedestrians, but also rare ones like strollers and animals for driving safety~\cite{kong2021opengan}.
A bio-image analysis system should recognize both commonly- and rarely-seen species for ecological research~\cite{romero2020improving, van2018inaturalist}.
This motivates the well-studied problem of long-tailed recognition (LTR), which trains on class-imbalanced data and aims to achieve high accuracy averaged across all the classes~\cite{zhang2021deep}.
LTR has attracted increasing attention especially using deep neural networks~\cite{cao2019learning, kang2019decoupling, yang2020rethinking}.

{\bf Status quo}.
Because common classes have significantly more training data than rare classes, they dominate the training loss, contribute the most of gradients, and obtain high accuracy~\cite{zhang2021deep}. 
Consequently, a naively trained model performs well on them but significantly worse on the rare classes (Fig.~\ref{fig:splashing}a).
The key to addressing LTR is to balance various aspects. 
Many methods propose to balance per-class data distributions during training by upsampling rare classes or downsampling common classes~\cite{chawla2002smote, estabrooks2004multiple, feng2021exploring}.
Some others balance the losses or gradients during training~\cite{tang2020long,  khan2017cost, cui2019class, cao2019learning}.
Some approaches adopt transfer learning that learn features on common classes and use the features to learn rare-class classifiers~\cite{wang2017learning, zhong2019unequal, liu2019large, jamal2020rethinking}. 
It shows that decoupling feature learning and classifier learning leads to significant improvement over models that train them jointly~\cite{kang2019decoupling}.
From benchmarking results, the state-of-the-art accuracy is achieved by either ensembling expert models~\cite{wang2020long, xiang2020learning, cai2021ace, guo2021long, feng2021exploring} or the adoption of self-supervised pretraining with aggressive data augmentation  techniques~\cite{cui2021parametric}.

\begin{figure*}[t]
\centering
{\small 
\bf How are per-class weight norms evolving in training ($x$-axis)? Classes are sorted w.r.t. cardinality ($y$-axis).} 
\vspace{0mm}
\\
{\bf \ \ }
    \hspace{3mm}
    {\small {\bf (a)} naive} \hspace{1.5cm} 
    {\small {\bf (b)} L2-normalization} \hspace{1.3cm} 
    {\small {\bf (c)} WD} \hspace{1.7cm} 
    {\small {\bf (d)} MaxNorm} \hspace{1cm} 
    {\small {\bf (e)} WD+MaxNorm} {\bf \ \ }\\
    \includegraphics[width=0.225\textwidth, clip, trim={2.1cm 0.5cm 2.3cm 2.85cm}]{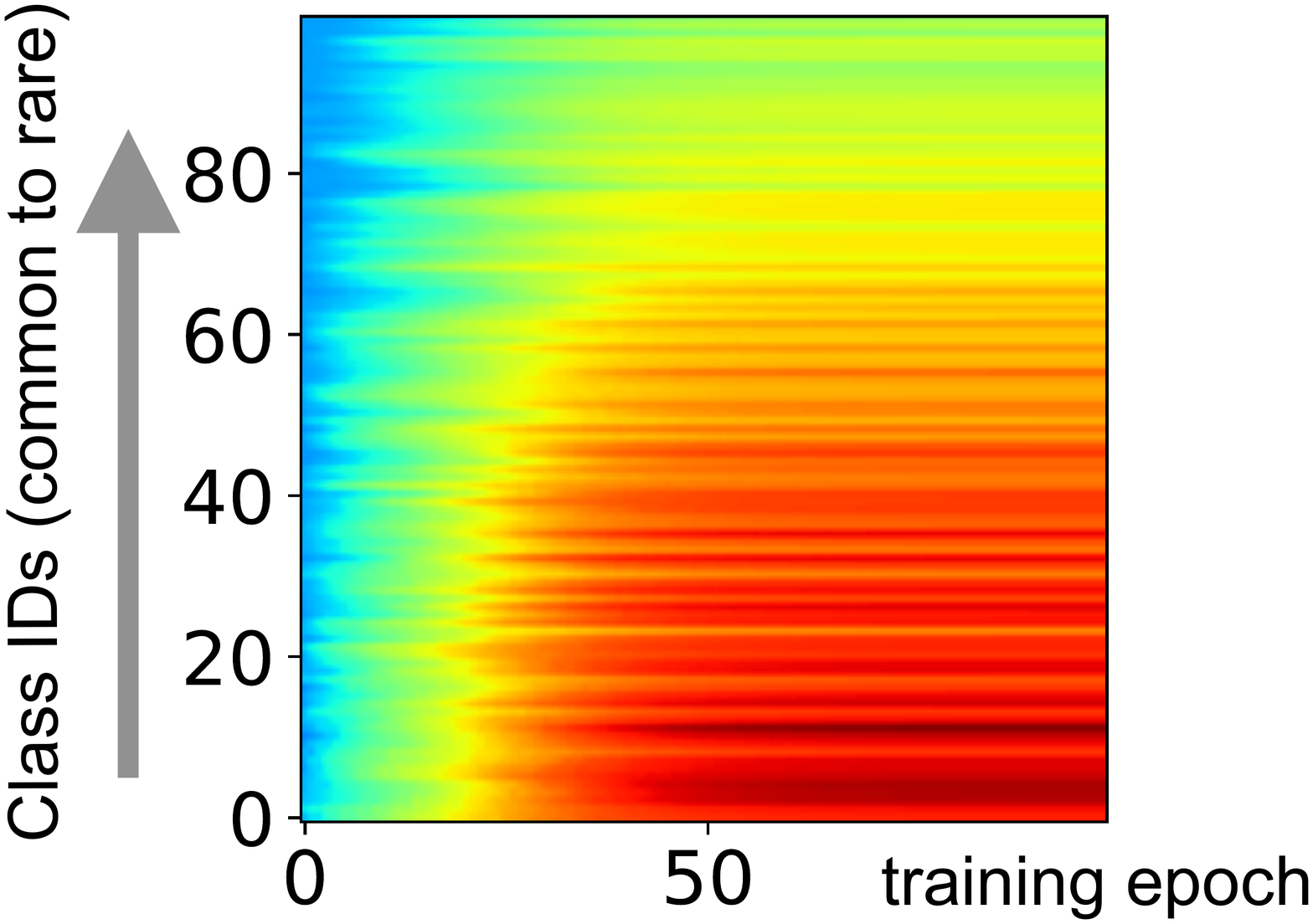}
    \includegraphics[width=0.18\textwidth, clip, trim={2.8cm 0cm 2.3cm 1cm}]{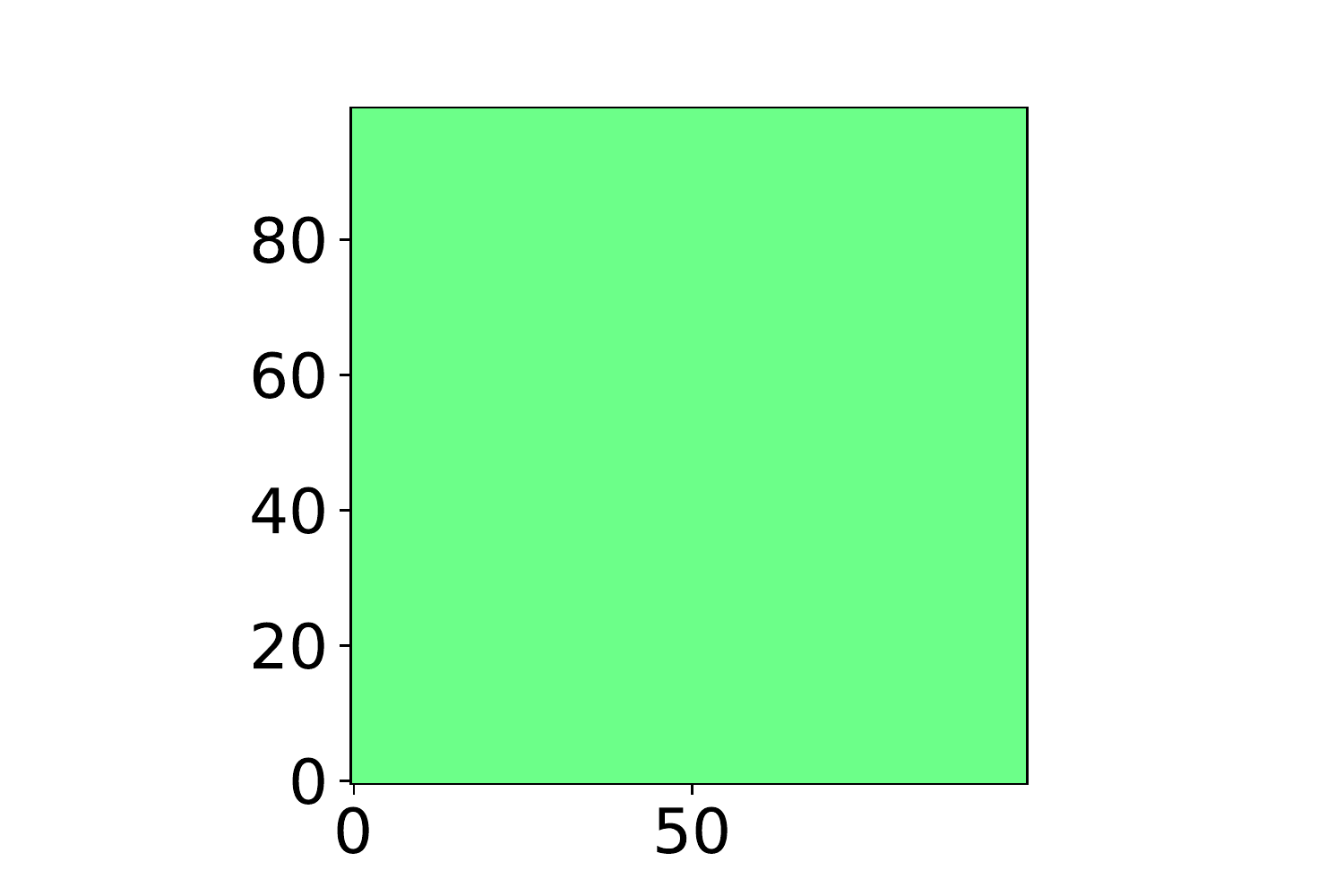}
    \includegraphics[width=0.18\textwidth, clip, trim={2.8cm 0cm 2.3cm 1cm}]{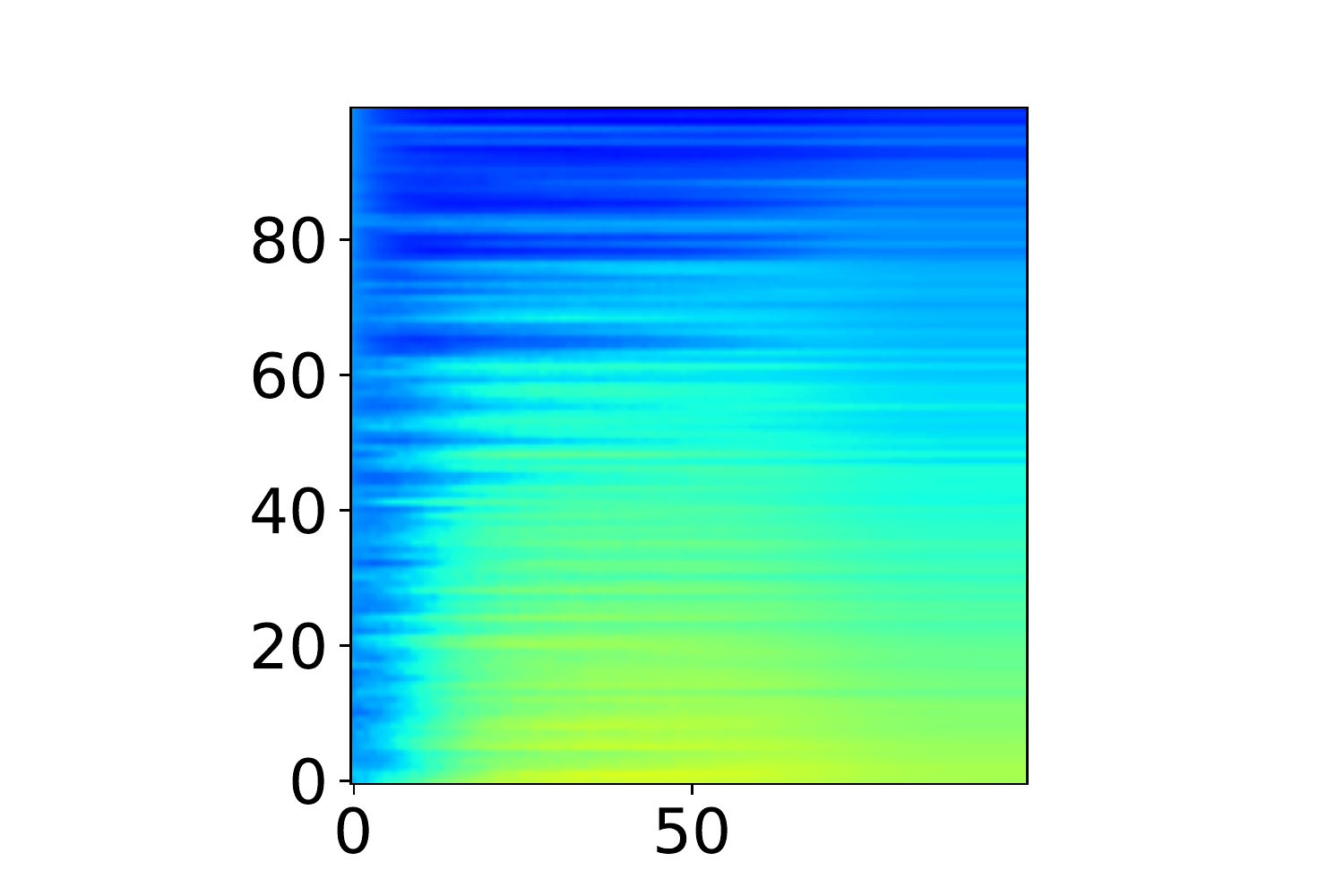}
    \includegraphics[width=0.18\textwidth, clip, trim={2.8cm 0cm 2.3cm 1cm}]{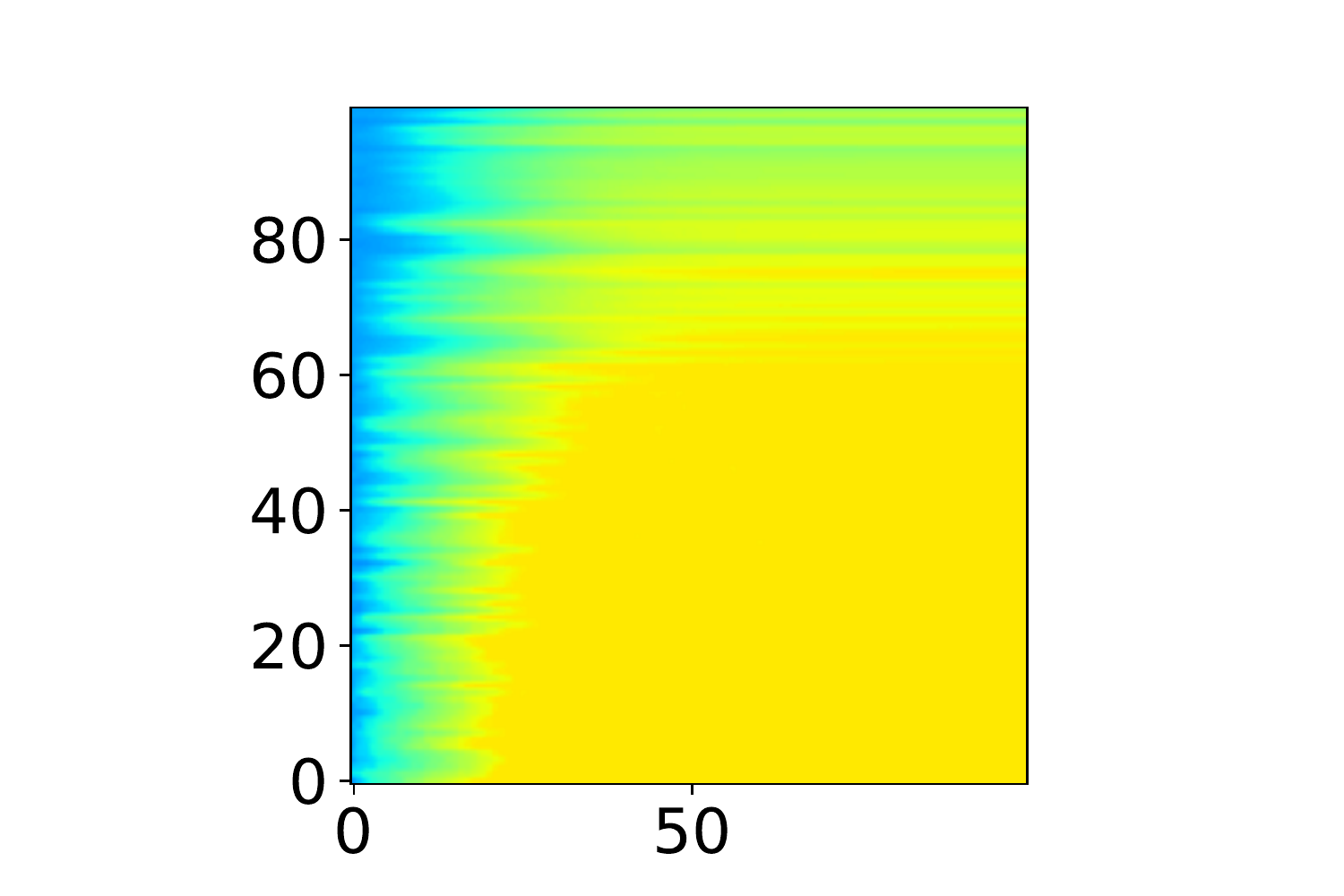}
    \includegraphics[width=0.20\textwidth, clip, trim={2.5cm 0cm 1.5cm 1cm}]{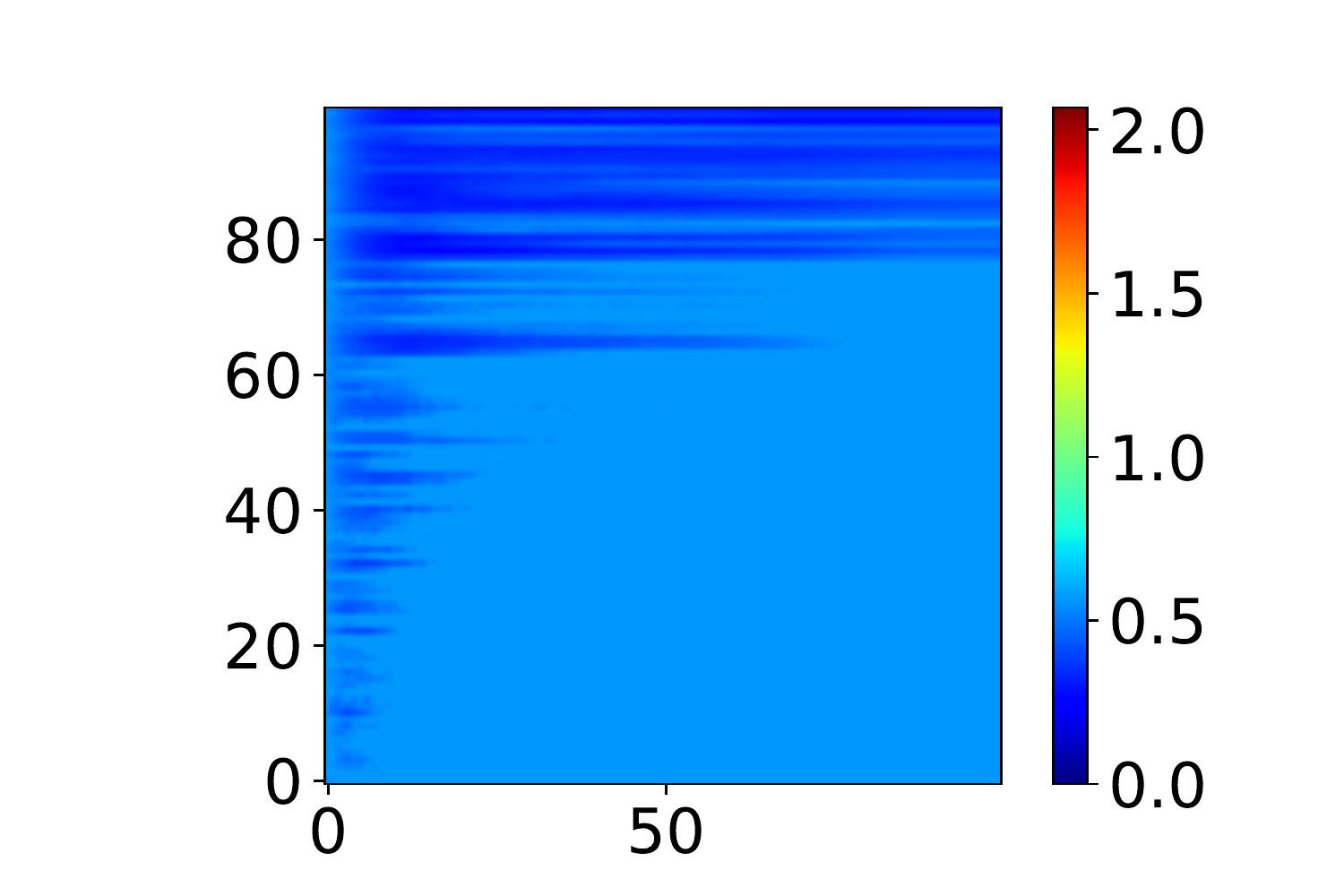}

    \vspace{-3mm}
\caption{\small
    {\bf (a)} In a naive model, all weights grow in norm during training, while those of common classes grow much faster.
    {\bf (b)} Because L2-normalization constrains weights to be unit-norm, weight norms stay constant during training.
    {\bf (c)} Weight decay (WD) regularizes all weights to be small while still allowing them to grow.
    {\bf (d)} MaxNorm caps large weights (of common classes) while letting small weights grow. 
    {\bf (e)} Combining weight decay and MaxNorm results in small 
    and balanced weights in norm.
    All plots share the same color map.
}
\label{fig:weightNormEvolving}
\vspace{-3mm}
\end{figure*}

{\bf Motivation}. 
We observe that a naively trained model on long-tailed class distributed data has ``artificially'' large weights for common classes (Fig~\ref{fig:splashing}b). Prior work also notes this observation~\cite{kang2019decoupling}.
Intuitively, this is because 
common classes have more training data that significantly grows classifier weights (Fig.~\ref{fig:weightNormEvolving}a).
This motivates our work to balance network weights  across classes for long-tailed recognition.
In contrast to existing methods (as exhaustively reviewed in a recent survey paper~\cite{zhang2021deep}), our work explores an orthogonal direction of  \emph{weight balancing}.

{\bf Contribution}.
To balance network weights in norm, we study three simple techniques.
We first point out that {\em L2-normalization} perfectly balances classifier weights to have unit norm (Fig.~\ref{fig:weightNormEvolving}b).
However, L2-normalization might be too strict to learn flexible parameters for better classifiers.
We then study \emph{weight decay}~\cite{hanson1988comparing, krogh1992simple} and the \emph{MaxNorm} constraint~\cite{shalev2011pegasos, hinton2012improving}.
Weight decay penalizes larger weights more heavily and so learns small balanced weights (Fig.~\ref{fig:weightNormEvolving}c); 
MaxNorm encourages growing small weights within a norm ball and caps all the weights by the radius (Fig.~\ref{fig:weightNormEvolving}d).
We find that both effectively learn balanced weights and boost LTR performance, although these well-known regularizers are  \emph{underexplored} in the LTR literature.
Please refer to Fig.~\ref{fig:splashing} for a nutshell of our work.

{\bf Key Findings}.
We show how simple regularizers boost LTR performance.
Without inventing new losses or adopting aggressive augmentation techniques or designing new network modules, 
we follow the simple two-stage training paradigm~\cite{kang2019decoupling} and derive a simple  approach that rivals or outperforms the state-of-the-art methods:  
(1) train a backbone using the standard cross-entropy loss by properly tuning \emph{weight decay}, and (2) train the classifier using a class-balanced loss by tuning \emph{weight decay} and \emph{MaxNorm}.
It is important to note how our simple approach  challenges the increasingly complicated LTR models, and hence serves as a strong future baseline for LTR.

\section{Related Work}
\label{sec:related_work}
 
{\bf Long-Tailed Recognition (LTR)}.
Real-world data tends to follow long-tailed class distributions, i.e., a few classes are commonly seen that have significantly more data than many classes that are infrequently / rarely seen.
As a result, a model naively trained on such data performs significantly worse on rare classes than common classes.
LTR requires training on such data to achieve high accuracy averaged
across all classes~\cite{cao2019learning, kang2019decoupling, yang2020rethinking}.
For LTR, numerous methods emphasize the accuracy on rare-classes. 
{\em Data re-balancing} techniques resample the training data to achieve a more balanced data distribution across classes~\cite{shen2016relay, mahajan2018exploring}, such as over-sampling rare-classes~\cite{chawla2002smote, han2005borderline} and undersampling common-classes~\cite{drummond2003c4}.
{\em Class-balanced loss reweighting} assigns weights to the classes~\cite{cui2019class, khan2017cost, cao2019learning, khan2019striking, huang2019deep, zhang2021distribution}, or even training examples~\cite{lin2017focal, shu2019meta, ren2018learning, khan2019striking}, aiming to modify their gradients to make the class-imbalanced data contribute properly to training.
{\em Transfer learning} methods transfer feature representations learned on the common-classes to the rare-classes~\cite{yin2019feature, liu2018open}.
Recent work examines the training procedure and finds LTR to be better addressed by decoupling feature learning and classifier learning, rather than training them jointly~\cite{kang2019decoupling, zhou2020bbn}. 
It is found that the SGD momentum causes issues in LTR that prevent further improvement~\cite{tang2020long} . 
Other sophisticated methods exploit self-supervised pretraining with more aggressive data augmentation techniques~\cite{cui2021parametric}, or ensemble expert models trained on different data regimes~\cite{cai2021ace, wang2020long}.
For a comprehensive review of the LTR literature, we refer the reader to the recent survey paper~\cite{zhang2021deep}.
Different from all the existing methods, we explore an orthogonal direction of {\em parameter regularization}, leading to a much simpler approach to LTR.


{\bf Parameter Regularization} adds extra information to solve an ill-posed problem, improving generalizability and preventing overfitting~\cite{ng2004feature, bishop2006pattern, buhlmann2011statistics}.
Regularization plays a crucial role in deep learning~\cite{krizhevsky2012imagenet}. One well-known  regularization is weight decay, which often applies L2-norm penalty on network weights~\cite{hanson1988comparing, krogh1992simple, loshchilov2017decoupled}. 
There exist many more regularizations~\cite{goodfellow2016regularization, kukavcka2017regularization}, such as weight normalization~\cite{salimans2016weight, ba2016layer},
MaxNorm constraints~\cite{shalev2011pegasos, hinton2012improving, goodfellow2016deep},
data augmentation~\cite{zhang2017mixup}, and dropout~\cite{hinton2012improving}.
In this work, we particularly examine the well-known yet {\em underexplored} regularizers in the LTR literature: L2-normalization, weight decay, and the MaxNorm constraint~\cite{shalev2011pegasos, hinton2012improving, goodfellow2016deep}.

{\bf Stage-wise Training} turns to be  effective in training deep networks~\cite{he2016deep, LoshchilovH17, yuan2018stagewise, zhao2020domain}. 
This can date back to stage-wise layer pretraining~\cite{hinton2006fast, bengio2007greedy}.
Recently, Kang et al. convincingly demonstrate that stage-wise training is important to LTR~\cite{kang2019decoupling}.
Concretely, Kang et al. propose to decouple feature learning and classifier learning into two independent stages~\cite{kang2019decoupling}: (1) feature learning using the standard cross-entropy loss, and (2) classifier learning over the learned feature using a  class-balancing loss.
While they did not explain why a single one-stage training with the class-balancing loss performed poorly, intuitively, this is 
because a class-balancing loss artificially scales up gradients computed from rare-class training data, which hurts the feature representation learning and hence the final LTR performance.
Follow-up work indirectly demonstrates this intuition with improved performance by stabilizing gradients during training~\cite{Ren2020balms, tang2020long}. 
In our paper, we adopt this two-stage training procedure, but focus on how to balance network weights for LTR. 

\section{Weight Balancing for Long-Tailed Learning}
\label{sec:weight-balancing}

{\bf Preliminaries}.
Long-tailed recognition (LTR) aims to train 
over a training set ${\cal D}$$=$$\{(\x_i, y_i)\}_{i=1}^N$, where data example $\x_i$ is labeled as $y_i$ $\in [1,\dots,K]$.
For class-$k$, ${\cal D}_k$ is the set of all its training examples and $|{\cal D}_k|$ is its cardinality.
The imbalance factor, $\text{IF}$$=$$\frac{\max_k|{\cal D}_k|}{\min_k|{\cal D}_k|}$,
measures how imbalanced the long-tailed training set is. For LTR, $\text{IF}$$\gg$$1$.
LTR emphasizes classification accuracy averaged
over classes,
i.e., accuracy=$\frac{1}{K}\sum_k${acc}$_k$, where acc$_k$ is the accuracy computed over testing examples of class-$k$.

LTR focuses on learning a $K$-way classification network $f(\cdot; \Tha)$ parameterized by $\Tha=\{\tha_{l, j}\}$, where $\tha_{l,j}$ is the $j^{th}$ filter weights at layer-$l$.
In a conv-layer, $\tha_{l,j}$ is a 3D kernel that convolves  the input (activation).
For brevity, we denote $\tha_k$ as the classifier filter corresponding to class-$k$.
Given a data example $\x_i$, the network predicts a label $y_i'=f(\x_i; \Tha)$. 
We measure the prediction error between $y'_i$ and the ground-truth $y_i$ using a cost function $\ell(y'_i, y_i)$, e.g., a cross-entropy (CE) loss~\cite{bishop2006pattern, murphy2012machine} or a class-balanced loss (CB)~\cite{cui2019class}. 
To train the network $f(\cdot; \Tha)$, we optimize $\Tha$ by minimizing $\ell(y'_i, y_i)$ over the whole training set $\cal D$:
\begin{align}
{\Tha^{*}} = \arg\min_{\Tha} F(\Tha; {\cal D}) \equiv\sum_{i=1}^N \ell\big(f(\x_i;\Tha), y_i\big).
\label{eq:obj}
\end{align}
Naively solving \eqref{eq:obj} produces a classifier (i.e., the last layer) that has artificially large weights in norm for common classes (Fig.~\ref{fig:splashing}b-left, Fig.~\ref{fig:weightNormEvolving}a). 
Therefore, we are motivated to learn a balanced classifier by regularizing classifier weights, denoted by $\tha_{k}$ for $k=1,\dots,K$.
Intermediate layers also have imbalanced filter weights (Fig.~\ref{fig:weightNorms_hiddenLayer}) even though a filter tends to fire on multiple classes~\cite{zeiler2014visualizing, netdissect2017}. Generally, one can also balance the weights at intermediate layers, and our study shows that doing so boosts performance.
Nevertheless, to simplify presentation in the following, we focus on regularization on the classifier weights $\tha_{k}$'s.

\subsection{Weight Balancing Techniques}
We examine the following three techniques to balance weights with respect to norms.


{\bf L2-normalization}.
A ``perfect'' way to balance the classifier weights $\tha_{k}$'s is to L2-normalize the classifier weights:
\begin{align}
{\Tha^{*}} = \arg\min_{\Tha} F(\Tha; {\cal D}), \quad  \text{\emph{s.t.} \ \ } \Vert \tha_{k} \Vert_2^2 = 1,  \ \ \forall \ k.
\label{eq:obj_L2norm}
\end{align}
As L2-normalization forces weights to be unit-length,
the classifier weights will have unit norm constant during training (Fig.~\ref{fig:weightNormEvolving}b).
Inspired by~\cite{kang2019decoupling},
we also post-hoc L2-normalize a trained classifier,
i.e.,
$\tha_{k}'$$=$$\tha_{k} / \Vert \tha_{k}\Vert_2$.
We find that post-hoc L2-normalization oftentimes improves LTR performance, favoring rare-classes yet sacrificing common-class accuracy. But it can also significantly decrease overall performance, e.g., on iNaturalist in Table~\ref{tab:other2datasets}.
Post-hoc L2-normalization is similar to the $\tau$-normalization~\cite{kang2019decoupling}, 
which allows varied per-class weight norms (rather than forcing them to be the same) and achieves better LTR performance.
This suggests that L2-normalization is too strict to strike a balance among the long-tailed distributed classes.
Importantly, our exploration finds that, while training with an L2-normalization constraint on the classifier improves over naive training, it underperforms the other two regularizers described below.

{\bf Weight Decay} is a well-studied technique~\cite{moody1991note, krogh1992simple} used to constrain a network by limiting the growth of the network weights. 
It decreases the complexity of the network, effectively mitigating overfitting and improving generalization.
Weight decay typically applies an L2-norm penalty to the network weights (we focus on the classifier $\tha_k$'s for now):
\begin{align}
{\Tha^{*}} = \arg\min_{\Tha} F(\Tha; {\cal D}) + \lambda \sum_{k}\Vert \tha_{k} \Vert_2^2,
\label{eq:WD_obj}
\end{align}
where $\lambda$ is a hyperparameter to control the impact of weight decay.
The weight decay term in~\eqref{eq:WD_obj} penalizes more heavily on large weights, preventing them from growing too large (Fig.~\ref{fig:weightNormEvolving}c)~\cite{moody1991note, krogh1992simple}. 
That said, weight decay encourages learning small balanced weights, as demonstrated by Fig.~\ref{fig:weightNormEvolving}.
Somewhat surprisingly, weight decay is underexplored in the literature of long-tailed recognition.
To the best of our knowledge, existing methods did not properly tune weight decay~\cite{cui2019class, tang2020long} (cf. code~\cite{code_LDAM, code_CBloss, code_Deconfound}) aside from their technical innovations.
This makes it unclear whether their improved LTR performance is due to better regularization inherent in these methods.
Importantly, our exploration demonstrates that, by simply tuning weight decay, we outperform most of the state-of-the-art methods on long-tailed benchmarks (Tables~\ref{tab:CIFAR100} and \ref{tab:other2datasets})!

{\bf MaxNorm Constraint}.
The third regularizer we explore is the MaxNorm constraint~\cite{shalev2011pegasos, hinton2012improving, goodfellow2016deep}.
MaxNorm caps weight norms within an L2-norm ball with radius $\delta$:
\begin{align}
{\Tha^{*}} = \arg\min_{\Tha} F(\Tha; {\cal D}), \quad  \text{\emph{s.t.} \ \ } \Vert \tha_{k} \Vert_2^2 \le \delta^2,  \ \ \forall \ k,
\label{eq:obj_maxNorm}
\end{align}
where the hyperparameter $\delta$ is the radius of the norm-ball.
Solving \eqref{eq:obj_maxNorm} can be efficiently done through  Projected Gradient Descent (PGD), which projects big weights that are outside the L2-norm ball onto the constraint set~\cite{shalev2011pegasos}. It simply applies a renormalization step after each batch update. 
Specifically, at each iteration, PGD first computes an updated $\tha_k$ and then  projects it onto the norm ball:
\begin{align}
\tha_{k} \leftarrow \min\Big(1, {\delta} / {\Vert\tha_{k} \Vert_2} \Big) * \tha_{k}.
\label{eq:PGD}
\end{align}
Different from L2-normalization that strictly sets the norm value for all the filter weights as 1, 
MaxNorm  relaxes this constraint that allows the weights to move within the norm-ball during training, as visualized in
Fig.~\ref{fig:weightNormEvolving}d.

\subsection{Further Discussion}
\label{ssec:discussion}

To better understand how and why the aforementioned regularizers work for long-tailed recognition, we discuss the following aspects.

\begin{figure}[t]
\centering
\includegraphics[width=0.48\textwidth]{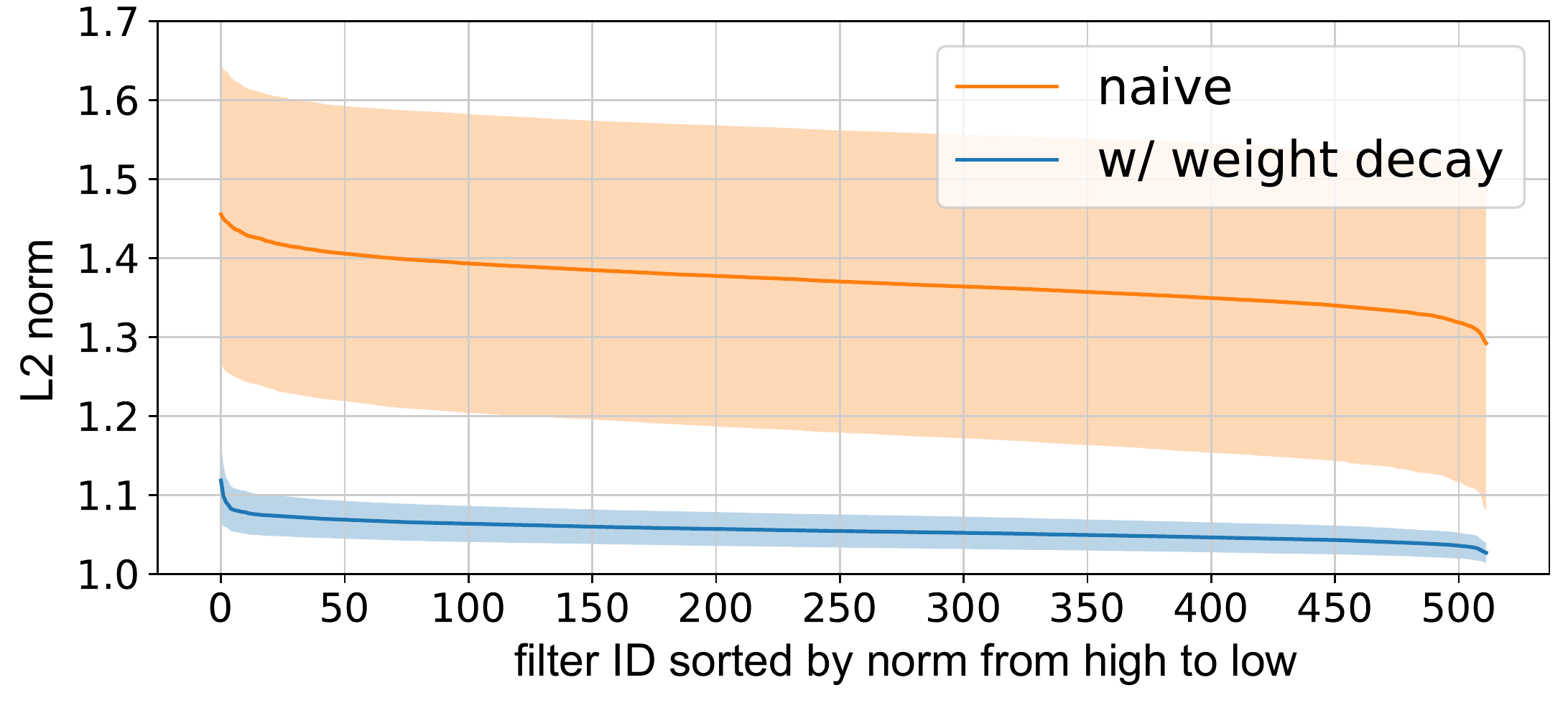}
\vspace{-8mm}
\caption{\small
{\bf Weight decay helps learn balanced weights at hidden layers}.
We compare the norm distribution at each layer (which has 512 filters) from the naive model (\orange{orange}) and the one trained with weight decay (\darkblue{blue}).
For each layer of a model, we sort the filter weights of each layer from high to low, compute their mean (the centerline) and variance (the shadow).
While individual filters in the hidden layers are not class-specific by design, recent work demonstrates that certain filters tend to fire on certain classes~\cite{zeiler2014visualizing, netdissect2017}; we still find them to be ``imbalanced'' in norms from the naive model. 
Weight decay encourages learning small and balanced filters, cf. its flat centerline and small variance.
}
\label{fig:weightNorms_hiddenLayer}
\vspace{-2mm}
\end{figure}

\begin{figure}[t]
\centering
\includegraphics[width=0.5\textwidth, clip, trim={4.5cm 0cm 4cm 0cm}]{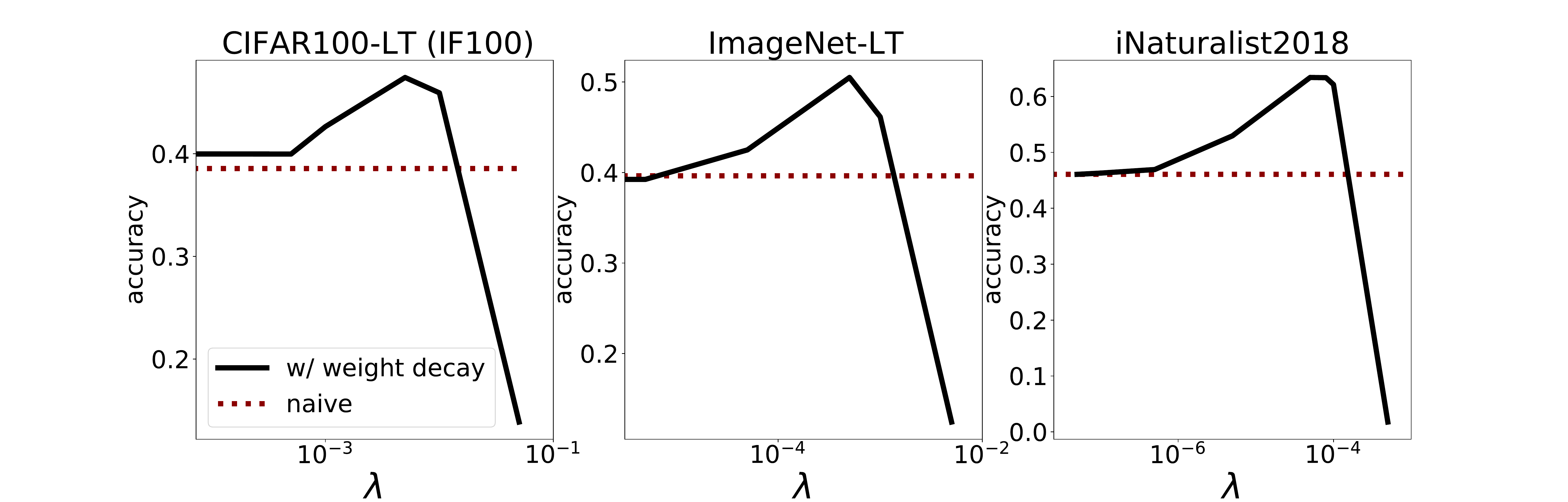}\\
\vspace{-3mm}
\caption{\small 
    Tuning weight decay drastically improves long-tailed recognition performance.
    We do not use any class-balancing techniques but simply use CE loss and tune weight decay $\lambda$ to regularize all network weight.
    For example, tuning $\lambda$ yields $46.1\%$ accuracy on CIFAR100-LT (IF=100), outperforming many state-of-the-art methods such as DiVE (45.4\%)~\cite{he2021distilling} and SSD (46.0\%)~\cite{li2021self}.
    By checking the publicly available code, we find that existing methods do not tune weight decay, e.g.,  \cite{cui2019class,cao2019learning} set $\lambda=$2e-4 (according to their code~\cite{code_LDAM, code_CBloss}), leading to poor accuracy 38.32\%.
    } 
\label{fig:perf_vs_wd}	
\vspace{-3mm}
\end{figure}

{\bf Weight Decay and MaxNorm}.
Both regularizers balance weight norms \emph{dynamically} during training, as opposed to L2-normalization which simply forces per-filter weights to be unit-length in norm.
Weight decay encourages learning small weights, and MaxNorm encourages weights to grow within a norm ball but cap them when their norms exceed the radius.
Weight decay pulls all weights to the origin. 
As a result, when $\lambda$ increases in \eqref{eq:WD_obj}, the weight decay penalty prevails $F(\Tha; {\cal D})$, making training unstable~\cite{bertsekas1976multiplier} (Fig.~\ref{fig:perf_vs_wd}). 
In contrast, MaxNorm does not pull weights towards the origin but simply caps the weight norms, and so has better numerical stability. 

Although weight decay and MaxNorm appear to be quite different, they are related that weight decay can be thought of as an immediate step when solving MaxNorm.
Let's rewrite the MaxNorm constrained objective function \eqref{eq:obj_maxNorm} by constructing a Lagrangian function:
\begin{align}
\Tha^* = \arg\min_{\Tha} \max_{\gamma \ge 0}  F(\Tha; {\cal D}) + \sum_{k} \gamma (\Vert \tha_{k} \Vert_2^2 - \delta),
\label{eq:obj_lagrangian_maxNorm}
\end{align}
where $\gamma$ is the Karush–Kuhn–Tucker (KKT) multiplier.
Suppose that we could solve  \eqref{eq:obj_lagrangian_maxNorm} using the coordinate descent method, i.e., iteratively optimizing over $\Tha$ and $\gamma$~\cite{wright2015coordinate}. 
When fixing $\gamma$, we have the same loss as \eqref{eq:WD_obj} which is constrained by weight decay, and $\gamma$ becomes the hyperparameter $\lambda$ to control weight decay.
That said, solving the weight decay constrained problem \eqref{eq:WD_obj} is a step of solving MaxNorm \eqref{eq:obj_maxNorm}.
Interestingly, we find that applying weight decay and MaxNorm jointly yields better performance than using each of them independently. 
This is probably because of their complementary advantages: (1) weight decay on the small weights still improves their generalization and reduces overfitting, and (2) MaxNorm prevents the large weights from dominating the training.


{\bf Extreme cases}. When $\delta$$\rightarrow$$\infty$ in MaxNorm, \eqref{eq:obj_maxNorm} boils down to the naive training \eqref{eq:obj}. 
On the other hand, a sufficiently small $\delta$ encourages all the weights to be close to the surface of the norm-ball. 
This is still different from the L2-normalization which strictly requires the weights to be on the surface. 
Compared to L2-normalization (Fig.~\ref{fig:weightNormEvolving}b), MaxNorm offers freespace within the norm ball to let weights grow (Fig.~\ref{fig:weightNormEvolving}d).
This intuitively explains why MaxNorm performs better than L2-normalization.

{\bf Weight decay can easily balance all network weights}.
We point out that weight decay regularizes classifier weights without the need to separate per-class filters.
This offers convenience in training, differently from MaxNorm which must separate each filter and scaling it w.r.t its norms.
Because of such a convenience, weight decay can be easily used to balance all network weights (Fig.~\ref{fig:weightNorms_hiddenLayer}).
In principle, MaxNorm can also be applied to all layers, but we find it non-trivial to do so, as this seems to require setting per-layer thresholds in \eqref{eq:obj_maxNorm} (tuning which is time-consuming).
While weight decay is widely used in network training,  we find that properly tuning it drastically improves long-tailed recognition accuracy (Table~\ref{tab:CIFAR100-ablation}).

\subsection{Training Pipeline}
\label{ssec:pipeline}

Because the aforementioned weight balancing techniques are not exclusive to each other, in principle, one can use a single technique or multiple ones together.
Recall that we follow the two-stage training paradigm~\cite{kang2019decoupling} in our work, which first trains a network for feature representation and then trains the classifier atop the learned features. 
This raises a question how to apply the weight balancing techniques effectively.
Among extensive exploration, we find that tuning $\lambda$ for weight decay in \eqref{eq:WD_obj} is sufficient to learn a generalizable feature representation as the first-stage training.
In contrast, applying MaxNorm is nontrivial because we find that it requires setting per-layer thresholds in \eqref{eq:obj_maxNorm}. This tuning process is time-consuming.
In the second-stage training (i.e., training the classifier), we find that tuning either/both weight decay and MaxNorm remarkably improves LTR accuracy.
Because the classifier training  simply involves only one layer (or two layers if we think of the top two as a non-linear classifier), tuning hyperparameters of the regularizers is quite efficient. To tune them, one can use random search~\cite{bergstra2012random} or Bayesian Optimization~\cite{snoek2012practical, Nogueira2014Bayesian}. We use the latter in this work.
In summary, our simple training pipeline consists of the following two stages:
\begin{enumerate}[noitemsep, topsep=0pt] 
\itemsep1pt
\item \emph{Feature learning}: train a network by using the cross-entropy loss and tuning weight decay.
\item \emph{Classifier learning}: train a classifier over the learned features using a class-balanced loss~\cite{cui2019class}, weight decay, and MaxNorm.
\end{enumerate}

\begin{figure}[t]
\centering
\includegraphics[width=0.48\textwidth]{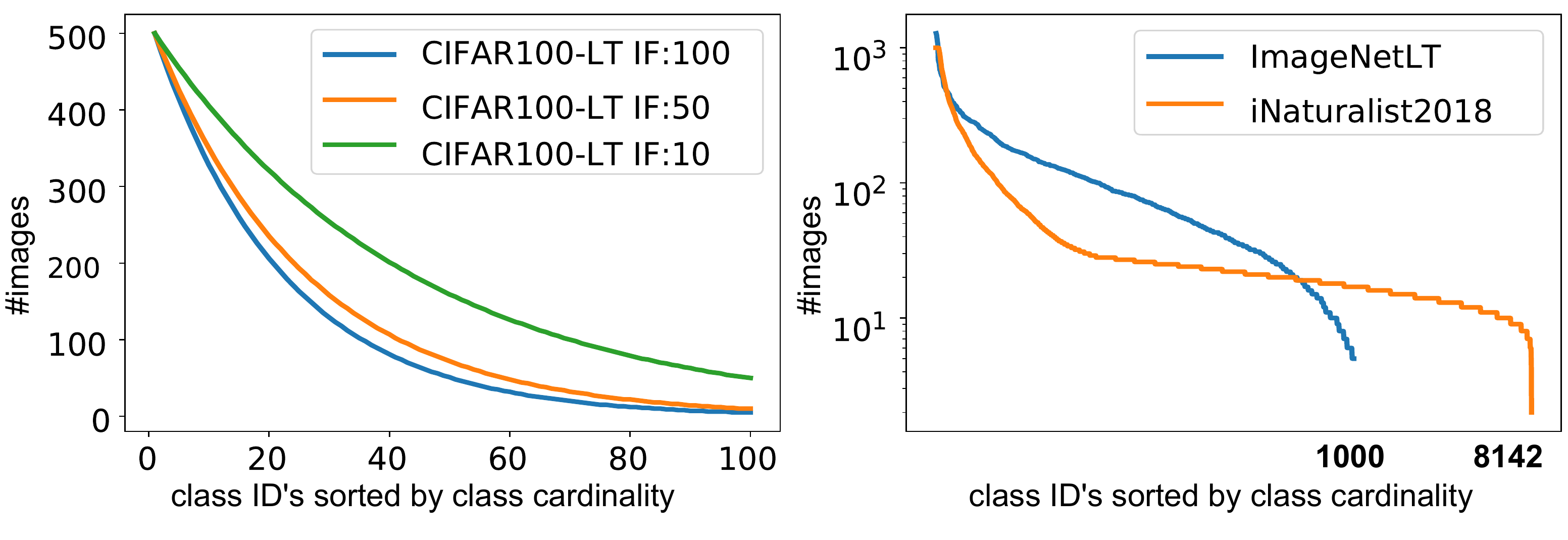}\\
\vspace{-3mm}
\caption{\small
Frequency distributions w.r.t class cardinality of five benchmarks.
{\bf Left:} We modify CIFAR100 by downsampling examples per class with different imbalance factors (IF) varying from 10 to 100.
{\bf Right:} We use two large-scale datasets: ImageNet-LT~\cite{liu2019large} that downsamples per-class images from ImageNet~\cite{deng2009imagenet},
and iNaturalist~\cite{van2018inaturalist} which is a real-world dataset with IF=500.
}
\label{fig:freq_datasets}
\vspace{-1mm}
\end{figure}


\section{Experiments}
\label{sec:exp}

We carry out extensive experiments to demonstrate how balancing network weights boosts long-tailed recognition performance.
First, we ablate the design choices in our pipeline as suggested in Section~\ref{ssec:pipeline}.
Then, we benchmark our methods on five established long-tailed datasets, showing that they rival or outperform existing LTR methods.
We start with the experiment setup.

\begin{figure*}[t]
\centering
\includegraphics[width=\textwidth, clip, trim={7cm 0cm 7cm 0cm}]{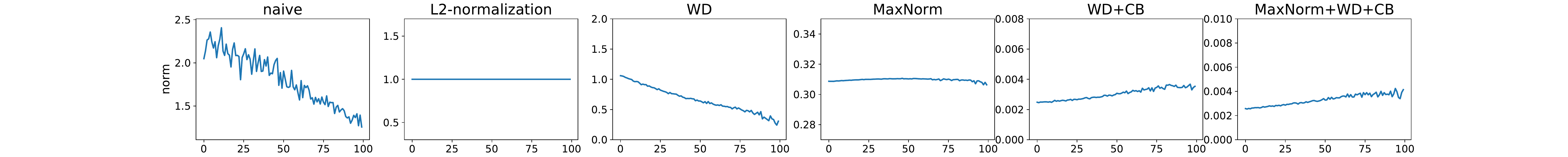}
\\
\includegraphics[width=\textwidth, clip, trim={7cm 0cm 7cm 0cm}]{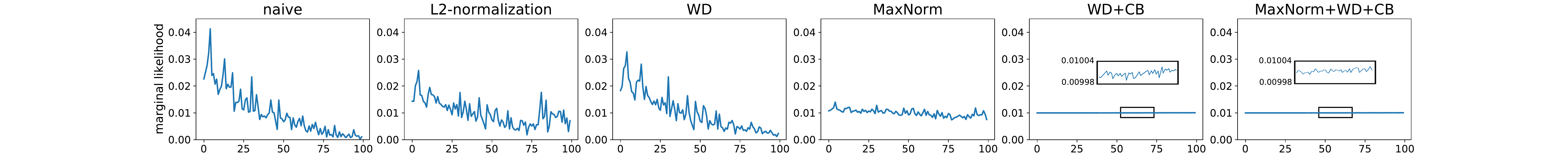}
\vspace{-8mm}
\caption{\small 
Per-class weight norms ({\bf top row}) and marginal likelihood ({\bf bottom row}) in the classifier vs. class ID sorted by class cardinality in decreasing order.
The plots are on CIFAR100-LT (IF100) val-set which has class-balanced data.
According to~\cite{Ren2020balms}, the ideal marginal likelihood should follow a uniform distribution.
Interestingly,  L2-normalization that ``perfectly'' balances weight norms does not produce ``uniform'' marginal likelihood. 
Weight decay (WD) slightly mitigates norm imbalance and marginal likelihood imbalance, but MaxNorm dramatically helps both.
The final model that incorporates MaxNorm, weight decay, and class-balanced loss yields nearly ``perfect'' marginal likelihood 
and balanced weights, which have a small bias towards rare-classes, presumably to emphasize their accuracy. 
} 
\label{fig:likelihood_distribution}	
\vspace{-1mm}
\end{figure*}

\subsection{Experiment Setup}
\label{ssec:exp_setup}

{\bf Datasets}. 
We use five long-tailed benchmarks.
Following \cite{cao2016deep}, we modify the CIFAR100 dataset~\cite{krizhevsky2009learning} by downsampling per-class training examples using some exponential decay functions, resulting in a long-tailed version, named \emph{CIFAR100-LT}. CIFAR100-LT still has 100 classes and a balanced validation set for evaluation.
By varying an imbalance factor (IF) $\in [100, 50, 10]$, we create three long-tailed training sets (Fig.~\ref{fig:freq_datasets}-left).
\emph{ImageNet-LT} is introduced in  \cite{liu2018open} by artificially truncating the balanced version  ImageNet~\cite{deng2009imagenet}. ImageNet-LT has 1,000 classes, and the number of per-class training data ranges from 5 to 1280.
\emph{iNaturalist2018}~\cite{van2018inaturalist} is a real-world dataset that has 8,142 naturally long-tailed classes. Fig.~\ref{fig:freq_datasets} summarizes the class frequency distributions of these datasets. 
ImageNet and iNaturalist2018 are publicly available for non-commercial research and educational purposes; CIFAR100 is  released under the MIT license.
We note that ImageNet and CIFAR100 have a ``people'' class or contain images that captured human faces and person signatures.
This is a concern related to fairness and privacy. Therefore, we cautiously proceed our research and \href{https://github.com/ShadeAlsha/LTR-weight-balancing}{release our code}
under the MIT License without re-distributing the data.


{\bf Network architectures}. For a fair comparison to prior art, we follow~\cite{cui2019class, kang2019decoupling, liu2019large, jamal2020rethinking, yang2020rethinking} to use specific network architectures on each dataset. 
We use ResNet32~\cite{he2016deep} on CIFAR100-LT, ResNeXt50~\cite{xie2017aggregated} on ImageNet-LT, and ResNet50~\cite{he2016deep} on iNaturalist2018. 


{\bf Evaluation protocol}. On each dataset, we train on the long-tailed class-imbalanced training set and evaluate on its (balanced) validation/test set. 
On ImageNet-LT, we tune hyperparameters and select models on its val-set and report performance on the test-set. 
On CIFAR100-LT and iNaturalist, which only have train-val sets,
we follow the literature~\cite{liu2019large} that uses the val-sets to tune and benchmark.
Following \cite{liu2019large}, we further report accuracy on three splits of classes that have varied numbers of training data: \emph{Many} ($>$100), \emph{Medium} (20$\sim$100), and \emph{Few} ($<20$).

{\bf Implementation}. 
We train our models using PyTorch toolbox~\cite{paszke2017automatic} on GeForce GTX 2080Ti GPUs.
The total time spent on this work is $\sim$2 GPU years with respect to this GPU type.
We train each  model for 200 epochs, with batch size as 64 (for CIFAR and ImageNet-LT) / 512 (for iNaturalist), SGD optimizer with momentum 0.9, and cosine learning rate scheduler~\cite{loshchilov2016sgdr} that gradually decays learning rates from 0.01 to 0. 
We also use random left-right flipping and cropping as our training augmentation.


\subsection{Ablation Study}
\label{ssec:ablation}

We study (1) the impact of weight decay in LTR, (2) how to regularize classifier learning, (3) classifier weight norms and marginal likelihood distribution, and (4) the evolution of weight norms during training. We use CIFAR100-LT (IF=100) for this study (unless stated otherwise).

{\bf Weight decay}.
We set a single constant $\lambda$ for all network parameters and focus on the first-stage training only, i.e., we use the standard cross-entropy loss to train a single network for classification.
Fig.~\ref{fig:perf_vs_wd} draws the top-1 accuracy as a function of $\lambda$ on the validation sets of three benchmarks.
Clearly, tuning $\lambda$ boosts accuracy, even outperforming many state-of-the-art methods (cf. Tables~\ref{tab:CIFAR100} and \ref{tab:other2datasets})!
Moreover, the optimal $\lambda$ varies for different datasets -- larger datasets need a smaller weight decay, intuitively because learning over more data helps generalization and so needs less regularization.

{\bf How to regularize classifier learning}.
To study how to apply the balancing techniques in the second-stage learning for classifiers, we also include  $\tau$-normalization~\cite{kang2019decoupling}  because it is an effective non-learned technique that post-hoc scales the classifier learned in the first stage.
We present salient conclusions based on the results in Fig.~\ref{fig:likelihood_distribution} (more in the supplement).
First, with an improved backbone (owing to a properly tuned weight decay in the first stage), $\tau$-norm boosts from 42.00\% to 51.31\%! 
This demonstrates the importance of learning a backbone that has balanced weights (Fig.~\ref{fig:weightNorms_hiddenLayer}).
Second, it is crucial to use a class-balanced (CB) loss~\cite{cui2019class} to learn the classifier. 
However, solely using the CB loss without regularizers only slightly improves (from 46.08\% to 47.09\%); once regularized with weight decay, it boosts to 52.42\%.
Third, applying both MaxNorm and weight decay improves further (53.35\%), and learning more layers (as a non-linear MLP classifier) improves to 53.55\%.


{\bf Classifier's weight norms and marginal likelihood}. 
Inspired by~\cite{Ren2020balms}, we examine the marginal likelihood based on predictions on the (balanced) test-set,
on which the ideal marginal likelihood follows a uniform distribution~\cite{Ren2020balms}.
We plot the marginal likelihood in Fig.~\ref{fig:likelihood_distribution}, alongside the norm distribution of different models.
Interestingly, L2-normalization that ``perfectly'' balances classifier weights does not produce balanced marginal likelihood.
In contrast, MaxNorm significantly helps  learn  balanced weights and balanced marginal likelihood. 
Combining MaxNorm, weight decay, and the CB loss, the model makes nearly ``perfect'' marginal likelihood with a small bias toward rare-classes in weight norms, presumably because it learns to emphasize rare-class accuracy.

{\bf Weight norm evolution during training}.
Fig.~\ref{fig:weightNormEvolving} depicts how classifier's weight norms evolve during training for different models. Briefly, without regularization, weights in the naive model grow fast in norm.
In contrast, weight decay prevents weights from growing too large, and MaxNorm quickly caps weights on a norm-ball surface and allows small weights to grow within the ball.

{
\setlength{\tabcolsep}{0.7em} 
\begin{table}[th]
\small
\centering		
\caption{\small 
{\bf Ablation study} on CIFAR100-LT (IF=100) w.r.t top-1 accuracy (\%).
``CE'': cross-entropy loss;
``CB'': class-balanced loss~\cite{cui2019class};
``WD'': weight decay;
``Max'': MaxNorm constraint;
``$\tau$-norm'': $\tau$-normalization~\cite{kang2019decoupling};
``+'': fine-tuning the last layer(s) as the second-stage training.
Here are salient conclusions.
(1) Learning with a properly tuned WD boosts performance from 38.38\% to 46.08\%, which is +8\% increase.
(2) Re-training the last layer with CB and WD gives another boost (+6\%) to 52.42\%.
(3) Based on the above, applying additional MaxNorm yields a slight improvement +1\% (53.35\%); finetuning the last two layers achieves 53.55\%.
(4) Finetuning more layers performs worse (cf. the supplement), presumably because CB induces modified gradients that affect feature learning and so hurt the final LTR performance.
}
\vspace{-3mm}
\begin{tabular}{lccccccc}		
\toprule
Model              & Many  & Medium & Few & All \\
\midrule
\multicolumn{5}{l}{\cellcolor{lightgrey}on the last layer (classifier)}\\
WD=0 (w/ CE)            &  64.05 & 35.80 & 11.43 & 38.38 \\
\ \ + $\tau$-norm       &  59.54 & 38.23 & 25.93 & 42.00 \\
WD tuned (w/ CE)        &  76.94 & 44.28 & 12.17 & 46.08 \\
\ \ + $\tau$-norm       &  73.11 & 47.69 & 30.10 & 51.31 \\
\ \ + L2norm            &  76.09 & 47.74 & 20.87 & 49.60  \\
\ \ + CE \& L2norm      &  76.37 & 48.11 & 21.00 & 49.87  \\
\ \ + CE \& WD          &  76.97 & 45.94 & 14.00 & 47.22 \\
\ \ + CB               &  {\bf 77.00} & 45.89 & 13.60 & 47.09 \\
\ \ + CB \& L2norm      &  76.43 & 48.20 & 21.60 & 50.10  \\
\ \ + CB \& WD          &  72.77 & 49.74 & 31.80 & 52.42 \\
\ \ + CB \& Max         &  76.49 & 49.23 & 20.67 & 50.20  \\
\ \ + CB \& WD \& Max   &  72.60 & {\bf 51.86} & 32.63 & 53.35  \\
\hline
\multicolumn{5}{l}{\cellcolor{lightgrey}on the last two layers}\\
\ \ + CB \& WD \& Max    &  71.37 & 51.17 & {\bf 35.53} & {\bf 53.55}  \\
\bottomrule		
\end{tabular}
\label{tab:CIFAR100-ablation}	
\vspace{-1mm}
\end{table}
}

{
\setlength{\tabcolsep}{1.2em} 
\begin{table}[th]
\small
\centering		
\caption{\small 
{\bf Benchmarking on CIFAR-100-LT} with different imbalance factors $[100, 50, 10]$ w.r.t top-1 accuracy (\%).
Please refer to the caption of Table~\ref{tab:CIFAR100-ablation} for abbreviations; ``+'' methods use CB loss.
WD makes a substantial impact on the training of LTR networks. Finetuning the classifier with proper regularization improves much further. This clearly shows the significance of parameter regularization in balancing weights for LTR.
Somewhat surprisingly, properly tuning weight decay in the two-stage training paradigm outperforms all existing methods on these three datasets. 
}
\vspace{-1mm}
\begin{tabular}{lccccccc}		
\toprule		
imbalance factor &  100		& 50   & 10 \\
\midrule		
CE~\cite{cui2019class} &  38.32  &  43.85  &  55.71\\
CE+CB \cite{cui2019class} &  39.60  &  45.32  &  57.99\\
KD \cite{hinton2015distilling} &  40.36  &  45.49  &  59.22\\
LDAM-DRW~\cite{cao2019learning} &  42.04  &  46.62  &  58.71\\
BBN~\cite{zhou2020bbn} &  42.56  &  47.02  &  59.12\\
LogitAjust \cite{menon2020long} &  42.01  &  47.03  &  57.74\\
LDAM+SSP~\cite{yang2020rethinking} &  43.43  &  47.11  &  58.91 \\ 
Focal~\cite{lin2017focal} & 38.41  &  44.32  & 55.78 \\ 
Focal+CB~\cite{cui2019class}  & 39.60  & 45.17  & 57.99 \\ 
De-confound~\cite{tang2020long} &  44.10   &  50.30   &  59.60 \\
$\tau$-norm~\cite{kang2019decoupling}  &  47.73  & 52.53  &  63.80 \\ 
SSD~\cite{li2021self}
                &  46.00  &  50.50  & 62.30 \\
DiVE~\cite{he2021distilling}            
                &  45.35  &  51.13  & 62.00 \\
DRO-LT~\cite{Samuel2021DistributionalRL}
                &  47.31  &  57.57  & 63.41 \\
PaCo~\cite{cui2021parametric} & 52.00 & 56.00  & 64.20\\
ACE (4-expert)~\cite{cai2021ace}  
                &  49.60  &  51.90  & ---  \\
RIDE (4-expert)~\cite{wang2020long}
                &  49.10  &  ---  &  ---  \\
\hline
\multicolumn{4}{c}{\cellcolor{lightlightgrey}{Our  methods (weight balancing)}} \\
naive           &  38.38  &  43.99  &  57.31 \\
WD              &  46.08  &  52.71  &  66.03 \\
\ \ + L2norm    &  49.60  &  56.33  &  67.16 \\
\ \ + $\tau$-norm&  51.31  &  57.65  &  67.79 \\
\ \ + WD        &  52.42  &  57.47  &  67.96 \\
\ \ + Max       &  50.24  &  56.06  &  67.10 \\
\ \ + WD \& Max &  {\bf 53.35}  &  {\bf 57.71}  &  {\bf 68.67} \\
\bottomrule		
\end{tabular}
\label{tab:CIFAR100}	
\vspace{-1mm}
\end{table}
}

\subsection{Benchmark Results}
\label{ssec:benchmark_results}

{\bf Compared Methods}. 
Considering the rapid evolution of the LTR field~\cite{zhang2021deep}, we compare against most relevant methods.
We choose methods that are recently published and representative of different types, such as Focal~\cite{lin2017focal} for loss reweighting, PaCo~\cite{cui2021parametric} for self-supervised pretraining and aggressive data augmentation, RIDE~\cite{wang2020long} for ensembling expert models, SSD~\cite{li2021self} and DiVE~\cite{he2021distilling} for transfer learning, etc.
For comparison, we report our methods including the naive model, the one trained with properly tuned weight decay, and models that have the second-stage learning for classifier with regularizers.
Tables~\ref{tab:CIFAR100} and \ref{tab:other2datasets} list benchmarking results on the CIFAR100-LT datasets, and ImageNet-LT and iNaturalist, respectively.

{\bf Results}.
Without bells and whistles, simply tuning weight decay (WD) in the first-stage training significantly boosts LTR performance over naive training and outperforms many prior methods.
For example, on CIFAR100-LT (IF100) in Table~\ref{tab:CIFAR100}, 
our WD model achieves 46.08\%, outperforming the naive model (38.38\%) and most of the compared methods including SSD (46.00\%)~\cite{li2021self} and DiVE (45.35\%)~\cite{he2021distilling}.
With the second stage (classifier learning), simply post-hoc modifying (without learning) the classifier (learned in the first stage) significantly improves performance from 46.09\% to 49.60\% (by L2-normalization) and to 51.31\% (by $\tau$-normalization).
By learning the classifier regularized with MaxNorm and/or weight decay, we achieve the state of the art (53.35\%).
Such a conclusion holds on all benchmarks.
However, on the two large-scale datasets ImageNet-LT and iNaturalists in Table~\ref{tab:other2datasets}, our methods rival prior art but underperform two types of methods that have ``bells and whistles'', including ensemble methods (RIDE~\cite{wang2020long} and ACE~\cite{li2021self}) that learn and fuse multiple models,
and self-supervised learning based methods (PaCo~\cite{cui2021parametric} and SSD~\cite{li2021self}) that adopt aggressive data augmentation techniques~\cite{he2020momentum, cubuk2020randaugment}.

{
\setlength{\tabcolsep}{0.15em} 
\begin{table}[t]
\small
\centering
\caption{\small
{\bf Benchmarking on ImageNet-LT and iNaturalists} in top-1 accuracy (\%).
Please refer to Table~\ref{tab:CIFAR100} for methods' names and salient conclusions.
We list the numbers of compared methods reported in their respective papers. 
Overall, our simple approach achieves competitive results with the prior methods particularly when they train a ``single (expert)'' model, although underperforms a few recent state-of-the-art methods which train and ensemble expert models (RIDE~\cite{wang2020long} and ACE~\cite{cai2021ace}), or adopt self-supervised pretraining (e.g., PaCo~\cite{cui2021parametric} and SSD~\cite{li2021self}) with aggressive data augmentation techniques~\cite{he2020momentum, cubuk2020randaugment}.
}
\vspace{-1mm}
\begin{tabular}{lcccc ccccc ccccc cc}
\toprule
 & \multicolumn{4}{c}{\tt ImageNet-LT} & & \multicolumn{4}{c}{\tt iNaturalist} \\	
\cmidrule(r){2-5} \cmidrule(r){7-10} 
 & Many & Med. & Few & All & & Many & Med. & Few & All\\
\midrule
CE~\cite{kang2019decoupling}  &  \underline{65.9}  &  37.5  &  7.7  &  44.4 
    &&  72.2  &  63.0  &  57.2  &  61.7\\
CE+CB \cite{cui2019class}  &  39.6  &  32.7  &  16.8  &  33.2
    &&  53.4  &  54.8  &  53.2  &  54.0\\
KD \cite{hinton2015distilling}  &   58.8  &  26.6  &  3.4  & 35.8 
    &&  \underline{72.6}  &  63.8  &  57.4  &  62.2 \\
Focal~\cite{cui2019class}  &  36.4  &  29.9  &  16.0  &  30.5
     &&  ---  &  ---  &  ---  &  61.1 \\
OLTR~\cite{liu2019large}  &  43.2  &  35.1  &  18.5  &  35.6
    &&  59.0  &  64.1  &  64.9  &  63.9 \\
LFME~\cite{xiang2020learning}  &  47.1  &  35.0  &  17.5  &  37.2
    &&  ---  &  ---  &  ---  &  --- \\ 
BBN~\cite{zhou2020bbn}  &  ---  &  ---  &  ---  &  ---
    &&  49.4  &  \underline{70.8}  & 65.3  &  66.3 \\
cRT~\cite{kang2019decoupling}  &  61.8 &  46.2  &  27.3  &  49.6
    &&  69.0  &  66.0  &  63.2  &  65.2 \\
$\tau$-norm~\cite{kang2019decoupling}  &  59.1  &  46.9  & 30.7  &  49.4
&&  65.6  &  65.3  &  65.5  &  65.6 \\
De-confound~\cite{tang2020long} 
                &   62.7  & 48.8    & 31.6    & 51.8
                &&  ---   & --- & --- & --- \\
DiVE~\cite{he2021distilling}            
                &  64.1 & \underline{50.4} & 31.5 & 53.1
                && 70.6 &  70.0 & 67.6 & 69.1 \\
DRO-LT~\cite{Samuel2021DistributionalRL}
                & 64.0 & 49.8 & 33.1 & 53.5
                && --- & --- & --- & 69.7 \\
DisAlign~\cite{zhang2021distribution} &  61.3  &  {\bf 52.2} & 31.4 & 52.9 && 69.0 & {\bf 71.1} & {\bf 70.2} &  {\bf 70.6} \\
\hline
\multicolumn{10}{c}{\cellcolor{lightlightgrey}{Our  methods (weight balancing) }} \\
naive   &  55.3  &  31.4  &  12.5  &  38.0
    && 54.7  &  46.0  &  43.9  &  46.1 \\
WD   &  {\bf 68.5}  &  42.4  &  14.2  &  48.6
    && {\bf 74.5}  &  66.5  &  61.5  &  65.4 \\ 

\ \ + L2norm   &  61.2  &  48.9  &  {\bf 42.6}  &  52.8
    && 11.2  &  47.4  &  66.9  &  51.3 \\
\ \ + $\tau$-norm   &  64.0  &  49.0  &  36.3  &  53.1
    && 71.3  &  69.8  &  68.9  &  69.6 \\

\ \ + WD   &  62.0  &  49.7  &  41.0  &  \underline{53.3}
    && 71.0	 &	70.3  &  69.4  & 70.0 \\
\ \ + Max   &  62.2  &  50.1  &  37.5  &  53.0
    && 71.4	& 68.9 & 69.1 & 69.2 \\
\ \ + WD \& Max   &  62.5  &  \underline{50.4}  &  \underline{41.5}  &  {\bf 53.9} 
    && 71.2 & 70.4 & \underline{69.7} & \underline{70.2} \\
\midrule
\midrule
\multicolumn{10}{c}{\cellcolor{lightlightgrey}{SOTA with ``bells and whistles'': ensembles, }} \\
\multicolumn{10}{c}{\cellcolor{lightlightgrey}{data augmentation, and self-supervised pretraining}} \\
RIDE~\cite{wang2020long}
                & 67.9 & 52.3 & 36.0 & 56.1  &&  66.5   & 72.1    & 71.5 &  71.3   \\
ACE~\cite{cai2021ace}  
                 &  ---  &  --- & --- & 56.6 && --- & --- & --- &  72.9  \\
SSD~\cite{li2021self}
                &  66.8  & 53.1 & 35.4 & 56.0 
                && --- & --- & --- &  71.5\\
PaCo~\cite{cui2021parametric} &  63.2  &  51.6  & 39.2 & 54.4 
&& 69.5 & 72.3 & 73.1 &  72.3 \\
\bottomrule
\end{tabular}
\label{tab:other2datasets}
\vspace{-1mm}
\end{table}
}

\section{Conclusion}
Long-tailed recognition (LTR) is a crucial challenge for real-world data that tends to be imbalanced.
Our work is motivated by the empirical observation that a model naively trained over long-tailed data has artificially large weights for common classes (because they have more data to train than rare classes).
We propose to learn balanced weights via parameter regularization, including weight decay and MaxNorm regularizers.
Our extensive study shows that properly applying these  regularizers greatly boosts LTR performance.
We introduce a simple approach that outperforms  prior art on five long-tailed benchmarks.
Because these regularizers are underexplored in the long-tailed literature, we hope our study draws attention from the practitioners that parameter regularization should be the first method to consider, when addressing real-world problems related to the long-tailed distribution.

{\bf Limitations}.
While we focus on the orthogonal direction of parameter regularization to address LTR, we have not studied how our approaches complement existing techniques. For example, how to balance weights in training each of the expert models, or how to balance the weights alongside sophisticated data augmentation and self-supervised pretraining. We also point out that other regularization techniques might be better at balancing weights, for example using L$p$-norm weight decay where $p$$\not=$2~\cite{benedek1961space}.
We leave them to future work.

{\bf Societal Impact}.
Because real-world data tends to follow long-tailed distributions, our work has multiple positive societal impacts. For example, addressing the long-tail proves an important direction for studying bias and fairness in recognition~\cite{NEURIPS2018_1f1baa5b}.
However, any system that makes it easier to train a fair classifier on long-tailed classes also makes it possible for a malicious agent to train a system that automatically discriminates against a certain subgroup for which only little training data is available. This is potentially a negative societal impact.

{\bf Acknowledgement}.
This work was supported by the CMU Argo AI Center for Autonomous Vehicle Research. SA was supported in part by the KAUST Gifted Student’s
Program (KGSP) and the CMU Robotics Institute Summer Scholars program. YXW was supported in part by NSF Grant 2106825 and the Jump ARCHES endowment.

\newpage
\appendix

\section*{}
\begin{center}
{\bf \large Appendix}
\end{center}
\emph{In the appendix, we first supplement the ablation study with more results to justify the use of regularizers for better learning for long-tailed recognition (LTR). 
We then present our open-source code in Jupyter Notebook as a self-explanatory tutorial.
Lastly, we attach a video demo that shows how weights change during training with different regularizers. 
}

\section{Detailed Ablation Study}

In Table~\ref{tab:CIFAR100-ablation-detailed}, we list more results in addition to the ablation study presented in the main paper.
Please refer to the caption for salient conclusions.

\section{Open-Source Code}
\label{sec:opensource}

{\bf Description}. 
We release our code with  two executable Jupyter Notebook files for demonstrating our approaches (w.r.t training and evaluation). 
The files will reproduce the results in the ablation study on the CIFAR100-LT dataset (with an imabalance factor 100).
The Jupyter Notebook files are sufficiently self-explanatory with detailed comments, and displayed output. The first file compares the first-stage training between naive training and training with weight decay.
The second file studies different regularizers in the second stage training.
We advise the reader to run the files in order (if running them) because the second stage training (i.e., the second demo file) requires the saved model by the first file. Running the first file takes $\sim$2 hours with a GPU (NVIDIA GeForce RTX 3090), and the second file takes a few minutes.

\begin{itemize}
\item \begin{verbatim}demo1_first-stage-training.ipynb\end{verbatim}  
    Running this file compares the first-stage training between a naive network (without weight decay) and a model with a tuned weight decay. It should achieve an overall accuracy $\sim$39\% and $\sim$46\% respectively on the CIFAR100-LT (imbalance factor 100).
\item \begin{verbatim}demo2_second-stage-training.ipynb\end{verbatim} 
    Running this file will compare various regularizers used in the second-stage training. It should achieve an overall accuracy $>$52\%.
\end{itemize}

{\bf Why Jupyter Notebook?}
We prefer to release the code using Jupyter Notebook (\url{https://jupyter.org}) because it allows for interactive demonstration for education purposes.
In case the reader would like to run python script, using the following command can convert a   Jupyter Notebook file {\tt XXX.ipynb} into a Python script:
\begin{verbatim}jupyter nbconvert --to script XXX.ipynb\end{verbatim}

{\bf Requirement}.
Running our code requires some common packages.
We installed Python and most packages through Anaconda. A few other packages might not be installed automatically, such as Pandas, torchvision, and PyTorch, which are required to run our code. Below are the versions of Python and PyTorch used in our work. 
\begin{itemize}
\item Python version: 3.7.4 [GCC 7.3.0]
\item PyTorch verion: 1.7.1
\end{itemize}
We suggest assigning $>$1GB space to run all the files. The code will save checkpoints after every training epoch.

{\bf License}.
We release open-source code under the MIT License to foster future research in this field.

{
\setlength{\tabcolsep}{0.45em} 
\begin{table}
\small
\centering		
\caption{\small 
{\bf Ablation study} on CIFAR100-LT (IF=100) w.r.t top-1 accuracy (\%).
``CE'': cross-entropy loss;
``CB'': class-balanced loss~\cite{cui2019class};
``WD'': weight decay;
``Max'': MaxNorm constraint;
``default-WD'': using the weight decay tuned for the first-stage training;
``$\tau$-norm'': $\tau$-normalization~\cite{kang2019decoupling};
``+'': fine-tuning the last layer(s) as the second-stage training.
Here are salient conclusions.
(1) Learning with a properly tuned WD boosts performance from 38.38\% to 46.08\%, that is +8\% increase.
(2) Re-training the last layer with CB and WD gives another boost (+6\%) to 52.42\%.
(3) Based on the above, applying additional MaxNorm yields a slight improvement +1\% (53.35\%); finetuning the last two layers achieves 53.55\%.
(4) Finetuning more layers performs worse, presumably because CB induces modified gradients that affect feature learning, and so hurt the final LTR performance.
}
\vspace{-3mm}
\begin{tabular}{lccccccc}		
\toprule
model              & many  & median & few & avg \\
\midrule
\multicolumn{5}{l}{\cellcolor{lightgrey}on the last layer (classifier)}\\
WD=0 (w/ CE)            &  64.05 & 35.80 & 11.43 & 38.38 \\
\ \ + $\tau$-norm ($\tau$ =1.0)  &  59.54 & 38.23 & 25.93 & 42.00 \\
WD tuned (w/ CE) &  76.94 & 44.28 & 12.17 & 46.08 \\
\ \ + $\tau$-norm ($\tau$ =1.9)  &  73.11 & 47.69 & 30.10 & 51.31 \\
\ \ + L2norm            &  76.09 & 47.74 & 20.87 & 49.60  \\
\ \ + CE \& L2norm      &  76.37 & 48.11 & 21.00 & 49.87  \\
\ \ + CE \& WD          &  {\bf 76.97} & 45.94 & 14.00 & 47.22 \\
\ \ + CE \& Max    &  76.80 & 47.26 & 15.10 & 47.95  \\
\ \ + CE \& Max \&  default-WD  &  76.89 & 47.06 & 13.90 & 47.55 \\
\ \ + CE  \& Max \& WD   &  76.80 & 47.51 & 14.40 & 47.83 \\

\ \ + CB               &  77.00 & 45.89 & 13.60 & 47.09 \\
\ \ + CB \& L2norm      &  76.43 & 48.20 & 21.60 & 50.10  \\
\ \ + CB \& WD          &  72.77 & 49.74 & 31.80 & 52.42 \\
\ \ + CB \& Max   &  76.49 & 49.23 & 20.67 & 50.20  \\
\ \ + CB \& Max \& default-WD & 76.20 & 48.91	& 21.50 & 50.24 \\

\ \ + CB \& WD \& Max   &  72.60 & {\bf 51.86} & 32.63 & 53.35  \\

\hline
\multicolumn{5}{l}{\cellcolor{lightgrey}on the last two layers}\\
\ \ + CE \& WD \& Max    &  76.34 & 48.46 & 21.17 & 50.03  \\
\ \ + CB \& WD \& Max    &  71.37 & 51.17 & {\bf 35.53} & {\bf 53.55}  \\

\hline
\multicolumn{5}{l}{\cellcolor{lightgrey}on the last five layers}\\
\ \ + CE \& WD \& Max    &  76.03 &	48.14 &	20.87 &	49.72  \\
\ \ + CB \& WD \& Max    &  74.37 &	49.80 &	26.63 &	51.45  \\

\bottomrule		
\end{tabular}
\label{tab:CIFAR100-ablation-detailed}	
\vspace{-3mm}
\end{table}
}

\section{Video Demo}
\label{lab:video}

\begin{figure*}[t]
\centering
{{\bf (a)} Naively trained network without weight decay} \\
\includegraphics[width=1\textwidth, clip, trim={5cm 0cm 5cm 0cm}]{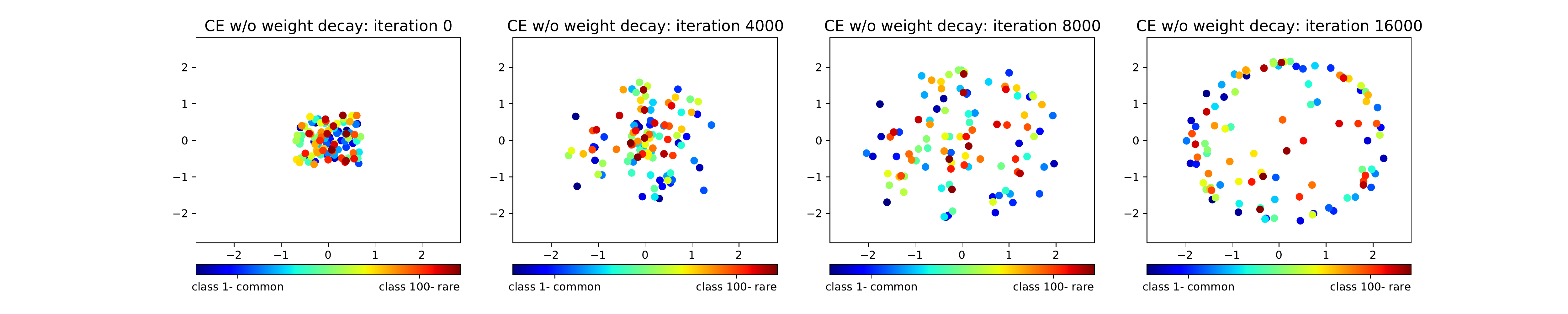}\\
\vspace{2mm}
{{\bf (b)} Network trained with L2-normalization} \\
\includegraphics[width=1\textwidth, clip, trim={5cm 0cm 5cm 0cm}]{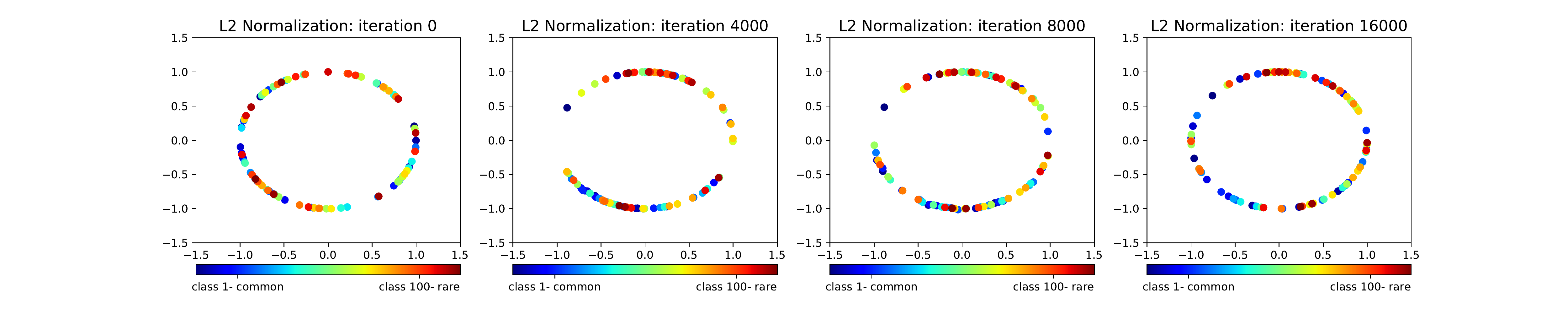}\\
\vspace{2mm}
{{\bf (c)} Network trained with weight decay} \\
\includegraphics[width=1\textwidth, clip, trim={5cm 0cm 5cm 0cm}]{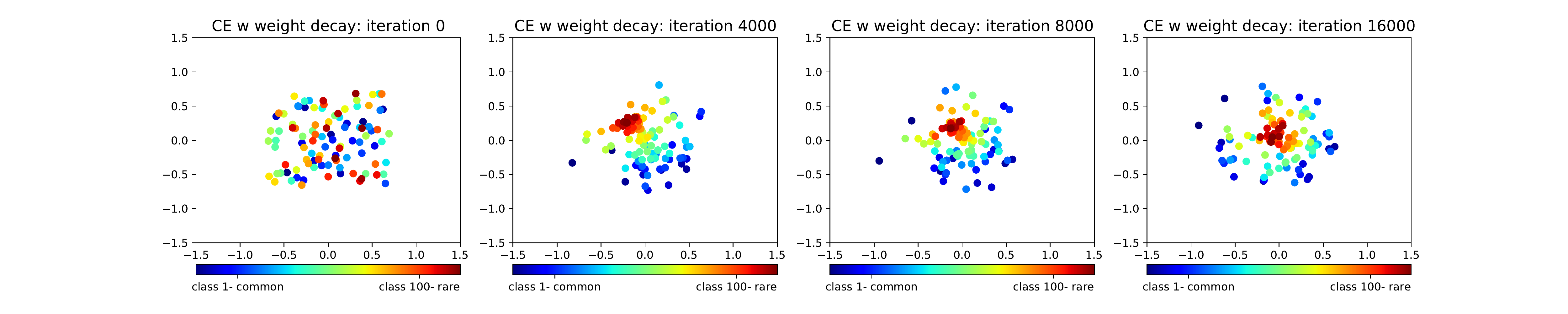}\\
\vspace{2mm}
{{\bf (d)} MaxNorm constrained network} \\
\includegraphics[width=1\textwidth, clip, trim={5cm 0cm 5cm 0cm}]{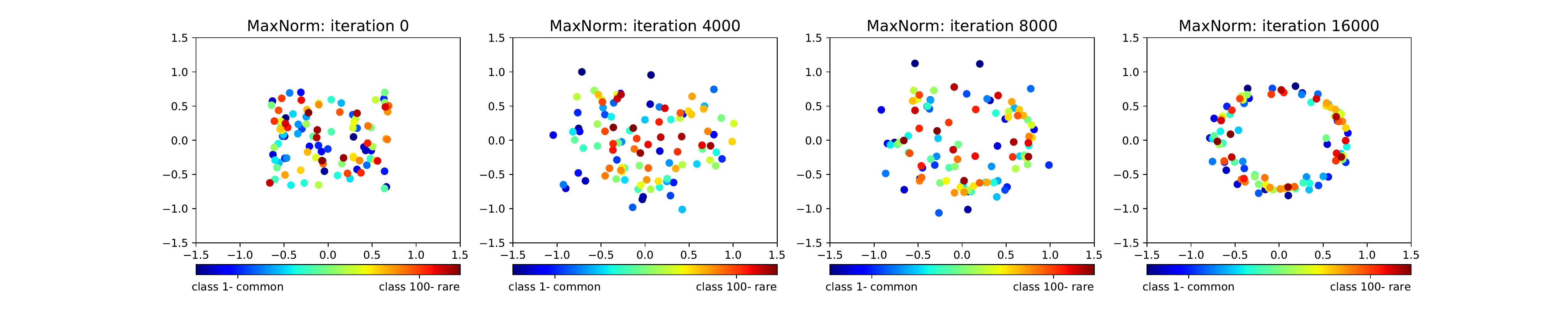}
\vspace{-4mm}
\caption{\small
We plot per-class filter weights of the classifier as 2D points from {\bf (a)} a naively trained network without weight decay,
{\bf (b)} a classifier with L2-normalization, 
{\bf (c)} a classifier with weight decay, 
and {\bf (d)} a classifier constrained by MaxNorm. 
All the networks are trained on the CIFAR100-LT dataset with imbalance factor as 100.
The four columns denote training iterations: iteration-0 as random initialization, iteration-4000, iteration-8000, and iteration-16000.
The naively trained network learns ``imbalanced'' weights, i.e., large weights for the common classes and small weights for the rare classes. The model trained with L2-normalization has constant weight norms.
When trained with weight decay, the network has smaller yet more balanced weights for all the classes.
The MaxNorm constrained network caps weight norms, encouraging small weights (from both common and rare classes) to grow, approaching to the surface of the norm ball.
We refer the reader to the video demo {\tt demo2D\_weight\_evolution.mp4} for better visualization.
} 
\label{fig:viz_dynamicRescaling}	
\end{figure*}

The goal of this section is to demonstrate how weights' norms  evolve during training. 
For demonstration, we train models on the CIFAR100-LT dataset with an imbalance factor 100.
To do so, we modify the ResNet34 network architecture by inserting an additional 2-dim pre-logit layer. 
This layer has weights $\W = [\w_{ij}] \in \RB^{2\times K}$ that project 2-dim pre-logit features to $K$-dim logits.
At the logit layer, each filter weight $\w_i$ (i.e., a row of $\W$) is class-specific.
Therefore, we can plot the $K$ 2-dim class-specific weights as $K$ points on a 2D plane.
The MaxNorm constraint upper bounds the norm of each class-specific weight, i.e., $\Vert \w_i\Vert_2<\delta$. 
Fig.~\ref{fig:viz_dynamicRescaling} plots per-class weights after three different training iterations.
For better visualization, we suggest the reader to watch our video demo {\tt demo2D\_weight\_evolution.mp4} in our github repository \url{https://github.com/ShadeAlsha/LTR-weight-balancing/blob/master/demo2D_weight_evolution.mp4}.

{\small
\bibliographystyle{ieee_fullname}
\bibliography{egbib_short}
}

\end{document}